\documentclass{article}

\PassOptionsToPackage{numbers,sort&compress}{natbib}
\usepackage[final,main]{neurips_2026}

\usepackage[utf8]{inputenc}
\usepackage[T1]{fontenc}
\usepackage{hyperref}
\usepackage{url}
\usepackage{booktabs}
\usepackage{amsfonts}
\usepackage{amsmath}
\usepackage{amssymb}
\usepackage{nicefrac}
\usepackage{microtype}
\usepackage{xcolor}
\usepackage{graphicx}
\usepackage{subcaption}
\usepackage{algorithm}
\usepackage{algorithmic}
\usepackage{multirow}
\usepackage{wrapfig}
\usepackage{float}
\usepackage[breakable,skins]{tcolorbox}

\setcitestyle{numbers}

\title{DetailAnywhere: Fashion Detail Generation via \\ Cross-Modal Feature Alignment Distillation}

\author{%
  \textbf{Zijun Li\textsuperscript{1,2}\thanks{Equal contribution.} \quad
  Yimin Zhou\textsuperscript{1}\footnotemark[1] \quad
  Jia Sun\textsuperscript{1}\thanks{Corresponding author.} \quad
  Honglie Wang\textsuperscript{1}} \\
  \textbf{Pengcheng Wei\textsuperscript{1} \quad
  Junlong Wu\textsuperscript{1} \quad
  Yongrui Heng\textsuperscript{1} \quad
  Jiyuan Wang\textsuperscript{1}} \\
  \textbf{Huan Ouyang\textsuperscript{1} \quad
  Boheng Zhang\textsuperscript{1} \quad
  Huaiqing Wang\textsuperscript{1} \quad
  Dewen Fan\textsuperscript{1}} \\
  \textbf{Qianqian Gan\textsuperscript{1} \quad
  Fan Yang\textsuperscript{1} \quad
 Tingting Gao\textsuperscript{1} }\\ 
  {\normalfont \textsuperscript{1}Kuaishou Technology}
  {\normalfont \textsuperscript{2}AIM for Health Lab, Monash University} \\
}

\begin{document}

\maketitle
\vspace{-2em}

\begin{figure}[H]
  \centering
  \includegraphics[width=0.9\linewidth]{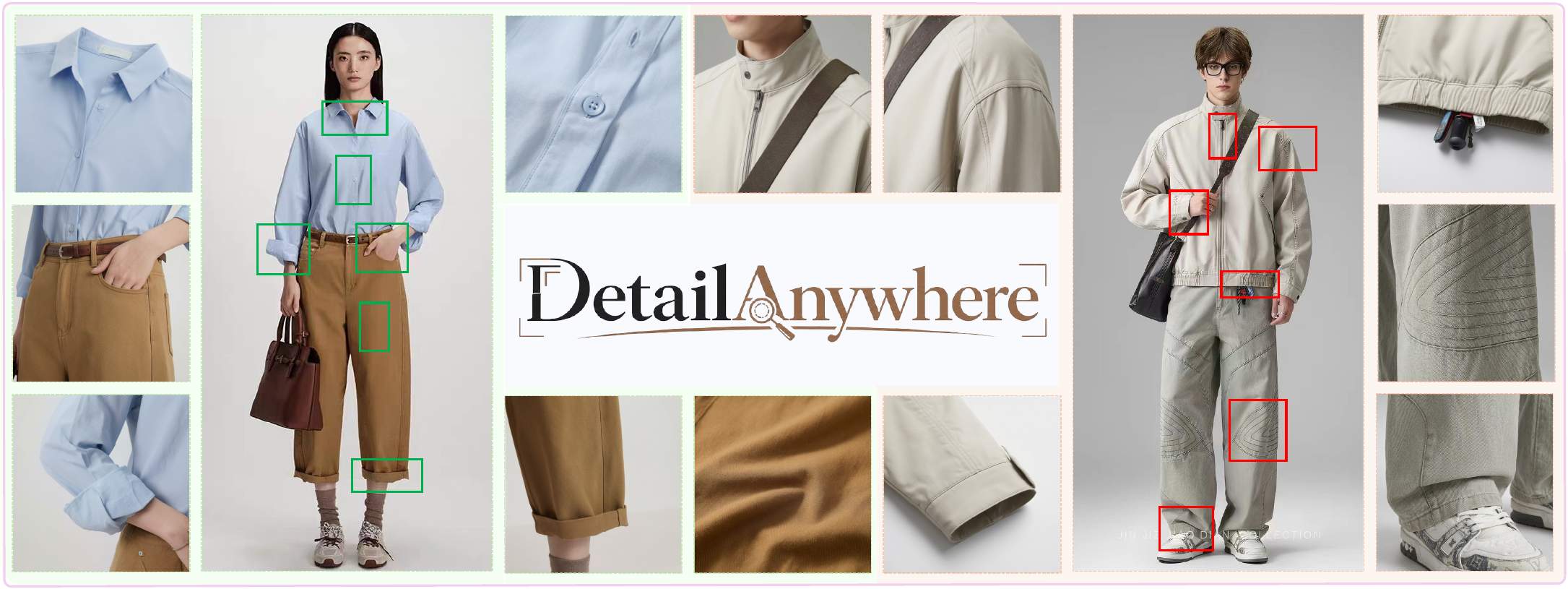}
  \caption{\textbf{Fashion Detail Generation.} Given a reference garment image with a bounding box indicating the target region, our model generates a high-fidelity, identity-consistent close-up of the indicated area.}
  \label{fig:teaser}
\end{figure}

\begin{abstract}
Diffusion-based generative AI has achieved remarkable success in e-commerce applications such as virtual try-on, poster generation, and product background synthesis.
However, when making online purchasing decisions for apparel, consumers also desire the freedom to examine specific detail regions of interest, such as collars, cuffs, and fabric textures, yet existing methods have not explicitly studied this setting.
We therefore formalize a new, non-template task: Fashion Detail Generation with focus conditioning, and release \textbf{FDBench}, the first benchmark comprising 40K+ human-verified reference-detail pairs across 41 different categories.
This task poses a unique semantic gap challenge: the model must bridge the correspondence between a focus marker on a product reference image and a photorealistic close-up view of the indicated region, while faithfully preserving the garment's identity, without any precise prompt.
To bridge this gap, we propose Cross-modal Feature Alignment Distillation (\textbf{CFAD}), which leverages a fine-tuned DINOv3 teacher to align both branches of a Multimodal Diffusion Transformer in a shared semantic space via dual-branch distillation.
To further improve consistency between generated details and reference images, we introduce a consistency reward model that jointly scores image pairs along three quality axes and optimizes generation via reinforcement learning.
Experiments show that our model \textbf{DetailAnywhere} significantly outperforms all state-of-the-art opensource methods across all metrics and human evaluations.
\end{abstract}

\section{Introduction}
\label{sec:intro}

Diffusion models~\cite{ho2020ddpm,rombach2022ldm,esser2024sd3} have become the dominant approach for high-fidelity image generation, and their impact on e-commerce visual content creation has been significant.
Virtual try-on systems~\cite{choi2024vitonhd, zhu2024tryondiffusion, choi2024idmvton} allow consumers to preview garment-body combinations without physical fitting rooms.
Automated poster generators~\cite{postermaker2025,chen2026posteromni} produce marketing materials with accurate text rendering and harmonious layouts.
Multimodal fashion editors~\cite{baldrati2023mgd,brooks2023instructpix2pix} support text- and sketch-guided garment manipulation.
These tools have dramatically lowered production costs while raising visual fidelity, and many have been deployed on major e-commerce platforms.

Despite this rapid progress, a critical gap remains: existing methods are essentially non-interactive from the consumer's perspective, delivering pre-designed templates or globally manipulated images with no mechanism for users to freely explore the specific regions they find most informative.
In practice, purchasing decisions for apparel frequently hinge on fine-grained garment details~\cite{li2025enhancing} that neither overall product images nor text descriptions can adequately convey~\cite{liu2024deepfashion,ecomiq2025}.
Consumers may wish to closely examine the stitching on a sweater cuff, the hardware finish of a zipper pull, or the weave pattern of a fabric---all of which directly shape perceived quality and drive purchasing confidence.
Such detail display images have therefore become an essential component of product demonstration.
However, acquiring them requires costly per-garment macro photography with specialized setups, imposing a heavy burden on small merchants and preventing scalable application.

Given the maturity of diffusion-based generation and the unified visual comprehension capabilities of recent multimodal models~\cite{qwenimage2025,labs2025fluxkontext}, a natural question arises: \textbf{can we generate high-quality detail views end-to-end from a single reference image?} We formalize this as the Fashion Detail Generation task, and introduce \textbf{FDBench}, which contains 40k paired reference and detail images.
Since providing a fine-grained text description that specifies both the product category and the target part (e.g., ``the seam-stitching details at the connection area between the hood brim of the anorak and the upper rear back panel'') is impractical for consumers, and the region of interest may lie at any location that lacks a standard part name, we define the focus marker as a bounding box that roughly delineates the area of interest.
As illustrated in Fig.~\ref{fig:teaser}, the expected output is a local detail image of the indicated region that faithfully preserves identity consistency with the reference while presenting the garment's fine-grained details in a clear and visually appealing manner.

However, generating detail views from a bounding-box-annotated reference presents two core challenges that distinguish it from standard image editing:
\textbf{(1)}~\textbf{Cross-branch semantic misalignment.}
State-of-the-art editing models adopt the MMDiT architecture~\cite{esser2024sd3,qwenimage2025,labs2025fluxkontext}, which fuses a text branch and an image branch through joint attention.
In Qwen-Image-Edit~\cite{qwenimage2025}, the reference image is encoded into the text branch by a vision-language encoder (Qwen2.5-VL), yet this branch alone cannot interpret the spatial semantics of a bounding-box marker, failing to convey what the box indicates to the image branch and leading to incoherent or off-target generation.
\textbf{(2)}~\textbf{Detail consistency degradation.}
Because our goal is to produce a close-up view of a specific local region, faithfully preserving the garment's identity (material, color, structural details) is essential.
However, existing image editing models~\cite{brooks2023instructpix2pix,ye2023ipadapter} tend to ``hallucinate'' fine-grained details that do not exist in the reference, especially when the target region receives insufficient attention during generation.

\begin{figure*}[t]
  \centering
  \includegraphics[width=\textwidth]{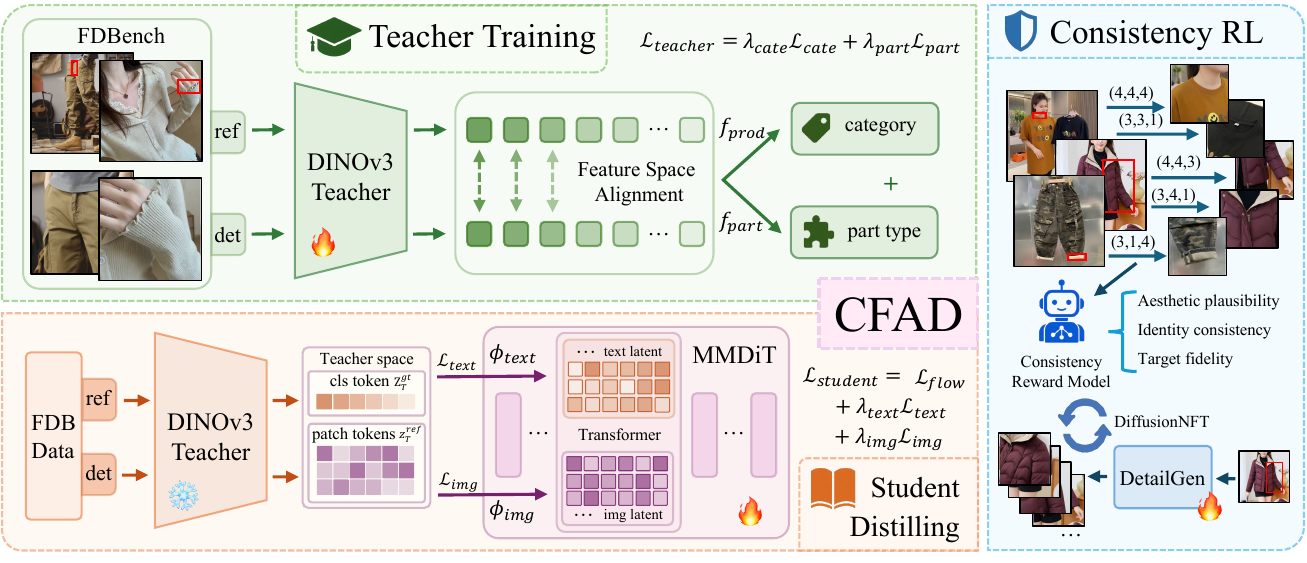}
  \caption{\textbf{Overview of DetailAnywhere.} Our framework consists of two stages: (1)~CFAD trains a view-bridging teacher to align reference and detail representations, then distills this alignment into both branches of the generator; (2)~A consistency reward model fine-tunes the generator via Negative-aware FineTuning to further improve identity preservation and detail fidelity.}
  \label{fig:method}
  \vspace{-1em}
\end{figure*}

Recent studies have shown that achieving cross-modal semantic alignment directly within large-scale generative models is notoriously difficult~\cite{chen2025multimodal, lee2025aligning, chefer2023attend}. However, REPA~\cite{yu2025repa} reveals that generative diffusion models implicitly learn internal representations aligned with those of small external visual encoders such as DINOv2~\cite{oquab2024dinov2,caron2021dino}.
This motivates a key insight: rather than forcing cross-modal alignment inside the large MMDiT, can we first complete the alignment in a small, trainable teacher model and then distill the feature distribution patterns back into the MMDiT?
Based on this idea, we propose \textbf{CFAD}, a dual-branch cross-modal feature alignment distillation framework, shown in Fig.~\ref{fig:method}.
We first train a view-bridging teacher model that maps both bounding-box-annotated references and detail close-ups into a shared semantic space, then distill this teacher's knowledge into both branches of the MMDiT~\cite{hinton2015distill}: the text branch receives semantic guidance from the teacher's encoding of the reference image, while the image branch receives structural guidance from the teacher's encoding of the ground-truth detail.
To tackle consistency degradation, we further train a consistency reward model and apply reinforcement learning via negative-aware finetuning~\cite{diffusionnft2025} to optimize generation along three complementary axes: aesthetic plausibility, identity consistency, and target fidelity.

In summary, our contributions are as follows:
\begin{enumerate}
    \item We formalize the fashion detail generation task and construct FDBench, the first benchmark for this problem, comprising $\sim$40K manually verified reference-detail pairs for training, and 800 pairs for evaluation. Experiments on FDBench demonstrate that our method achieves state-of-the-art results across all evaluation metrics.
    \item We propose CFAD (Cross-modal Feature Alignment Distillation), which offloads the difficult cross-modal alignment within a large-scale generative model to a small, easily-trained discriminative teacher, then distills the aligned representations back into both branches of the MMDiT, bypassing direct branch-to-branch optimization.
    \item We introduce a consistency reward model that scores reference-detail pairs along three task-specific quality dimensions and fine-tunes the generator via reinforcement learning, further improving identity consistency.

\end{enumerate}

\section{Related Work}
\label{sec:related}

\subsection{Image Generation and Editing}

Diffusion models~\cite{ho2020ddpm,rombach2022ldm} have become the dominant paradigm for image synthesis.
Early advances established key guidance mechanisms: classifier guidance~\cite{dhariwal2021diffusion} and classifier-free guidance (CFG)~\cite{ho2022cfg}, which significantly improved controllability and sample fidelity.
In parallel, CLIP~\cite{radford2021clip} enabled language-aligned visual supervision and became a core signal for text-guided generation and editing.
Latent diffusion models (LDMs)~\cite{rombach2022ldm} then improved efficiency by operating in a compressed latent space, while transformer-based variants, including DiT~\cite{peebles2023dit} and MMDiT~\cite{esser2024sd3}, further scaled quality by jointly processing text and image tokens through shared attention.
Flow matching~\cite{lipman2023flow} provides an efficient training framework for these architectures.
Subsequent works have further expanded diffusion-based modeling to a broad range of applications and explorations~\cite{wang2026geometry,wang2026learning,wang2025template,gong2025sculpting,wang2025editor,yan2025symmcompletion,xia2025diffpc}.

For image editing, InstructPix2Pix~\cite{brooks2023instructpix2pix} enables natural-language-guided editing, while ControlNet~\cite{zhang2023controlnet} and IP-Adapter~\cite{ye2023ipadapter} introduce spatial and image-prompt conditioning.
SDEdit~\cite{meng2022sdedit} supports stochastic editing through forward-backward diffusion.
Diffusion-based models and representations have also been increasingly applied to other domains such as 3D vision~\cite{li2025diffpcn,wang2025jasmine,yan2025posemaster,Wang_2024,Wang2_2024}.
REPA~\cite{yu2025repa} shows that diffusion transformers implicitly build discriminative representations similar to those of pretrained encoders like DINOv2~\cite{oquab2024dinov2,caron2021dino}, and that explicitly aligning the two accelerates training and improves quality.
DDAE~\cite{xiang2023ddae} and DIFT~\cite{tang2023dift} further demonstrate that diffusion models can act as unified self-supervised learners whose internal features encode fine-grained spatial correspondence.
Knowledge distillation~\cite{hinton2015distill} and cross-modal distillation~\cite{gupta2016crossmodal} provide the broader foundation for such representation transfer.
However, all the above methods target global edits or localized modifications within the same viewpoint.
Our task requires generating a novel close-up view of a specified local region, which demands cross-view semantic understanding that existing editing pipelines do not address.

\subsection{E-commerce Content Generation}

AI-assisted content creation for e-commerce has been an active area of research.
Existing studies can be grouped into several representative directions: virtual try-on~\cite{choi2024vitonhd, kim2024stableviton, chong2025catvton, morelli2023ladivton, yang2024tpd}, fashion image editing with specific controls~\cite{baldrati2023mgd,brooks2023instructpix2pix,ye2023ipadapter}, product-background and scene synthesis for merchandising, and automatic poster generation for marketing content~\cite{postermaker2025,chen2026posteromni}.

Despite this breadth, most pipelines operate at the global image level and do not model fine-grained local detail generation under user-specified spatial focus.
Our work addresses this gap by targeting identity-consistent close-up generation from a single reference image and a bounding box.
\subsection{Reinforcement Learning for Image Editing}

Reinforcement learning (RL) has been applied to align diffusion models with human preferences.
DDPO~\cite{black2024ddpo} formulates denoising as a multi-step RL problem; DPOK~\cite{fan2024dpok} adds KL regularization for stability.
GRPO~\cite{shao2024deepseekgrpo} offers group-relative policy optimization without a critic network, while DiffusionNFT~\cite{diffusionnft2025} introduces a forward-process RL formulation with continuous timestep support.
To support RL training, several reward models have been developed:
ImageReward~\cite{xu2024imagereward} and HPSv2~\cite{wu2023hpsv2} provide general-purpose preference scores for text-to-image generation, while EditReward~\cite{wu2025editreward} and EditScore~\cite{luo2025editscore} focus specifically on the quality of instruction-guided image edits.
In the e-commerce domain, business-oriented signals such as CTR prediction~\cite{guo2017deepfm,fan2025autoppautomatedproductposter} and conversion-rate feedback have also been explored, reflecting a trend toward task-specific, domain-grounded rewards.
RL-based or reward-driven optimization has also been adopted for multimodal visual reasoning, video reward modeling, and self-evolving MLLMs~\cite{jiang2026vlm,wu2025arc,wang2026cac,zhou2026high,heng2026eve}.

However, none of these reward models captures the fine-grained consistency requirements of fashion detail generation.
We therefore design a task-specific consistency reward model along three complementary dimensions and use it to guide generator fine-tuning via RL.

\section{Method}
\label{sec:method}

\subsection{Task Formulation}

We define the fashion detail generation task as follows:
given a reference garment image $\mathbf{x}_{\text{ref}} \in \mathbb{R}^{H \times W \times 3}$ depicting a full-body or half-body view of a clothing item, together with a bounding box $b$ rendered directly on the image to indicate the specified target region, the goal is to generate a detail image $\mathbf{x}_{\text{det}} \in \mathbb{R}^{H' \times W' \times 3}$ that presents a high-fidelity close-up of that region.

Formally, we seek to learn a conditional generator $G_\theta$ such that:
\begin{equation}
    \mathbf{x}_{\text{det}} = G_\theta(\mathbf{x}_{\text{ref}}^{\text{box}}), \quad \mathbf{x}_{\text{ref}}^{\text{box}} = \text{render}(\mathbf{x}_{\text{ref}}, b)
\end{equation}
where the generated detail must satisfy two first-class objectives:
\textbf{(i)} High fidelity: photorealistic and visually pleasing rendering of the local region, free of artifacts;
\textbf{(ii)} High consistency: faithful preservation of the garment's material, texture, color, and structural identity relative to the source.

\subsection{CFAD: Cross-Modal Feature Alignment Distillation}
\label{sec:cfad}

\paragraph{The text-branch bottleneck and prompt enhancement.}
Leading image generation and editing models such as SD3~\cite{esser2024sd3}, FLUX.1~\cite{flux2024}, and Qwen-Image-Edit~\cite{qwenimage2025} all adopt the MMDiT architecture, which fuses a text branch and an image branch through joint attention in each transformer block.
When applied to our task, these models struggle to associate the bounding-box marker on the reference with the target local region, producing incoherent or off-target detail images~\cite{chen2025multimodal, kim2024textembedding}.
The fundamental difficulty is that the text branch encodes a holistic understanding of the reference image but lacks region-level semantic grounding. It cannot convey which specific region the box indicates to the image branch.
We observe empirically that employing a large vision-language model to localize and describe the boxed region, then injecting this description into the editing prompt (denoted ``PE'' in Table~\ref{tab:main_results}), substantially improves baseline performance, confirming that enriching the text branch with region-level semantics is a key factor in bridging the cross-branch gap.
However, textual descriptions inherently cannot fully convey fine-grained structural cues and texture characteristics of the target region. Moreover, this approach introduces an expensive external VLM at inference time.

\paragraph{Key insight: feature-level alignment via a discriminative teacher.}
Recent work reveals that generative diffusion transformers already possess rich semantic representations internally.
REPA~\cite{yu2025repa} and follow-ups~\cite{leng2025repa,zheng2025diffusion} show that intermediate features of diffusion models closely resemble those of pretrained discriminative encoders such as DINOv2~\cite{oquab2024dinov2,caron2021dino}, and that explicit alignment accelerates training.
Complementarily, DDAE~\cite{xiang2023ddae} and DIFT~\cite{tang2023dift} demonstrate that diffusion models can serve as unified representation learners whose internal activations encode fine-grained spatial correspondence.
Together, these findings imply that the MMDiT's text and image branches already possess the \emph{latent capacity} for rich semantic understanding. What they lack is not representational power, but an explicit mechanism to align their internal features into a shared space where joint attention can effectively bridge the cross-view correspondence.

This motivates a simple yet effective idea: rather than injecting region understanding through PE, we can implicitly align the text and image branches in a shared feature space, maximally leveraging the powerful representational capacity that the DiT already possesses.
Specifically, we train a small discriminative teacher that accepts both the bounding-box-annotated reference and the corresponding detail close-up, and maps them into a shared semantic space via classification training.
Following and extending the REPA paradigm~\cite{yu2025repa}, we then distill these aligned representations into both branches of the MMDiT simultaneously, which we call Cross-Modal Feature Alignment Distillation (CFAD).
Because the two distillation targets reside in the same teacher space by construction, the text and image branches are implicitly guided toward mutual alignment through the teacher as an intermediary, entirely bypassing both the language bottleneck of PE and the intractable direct branch-to-branch optimization within the large generative model.

\paragraph{View-bridging teacher learning.}
We realize CFAD through two stages.
In the first stage, we train a lightweight discriminative teacher that unifies the representation space of both views, providing the aligned targets for subsequent distillation.

Specifically, we fine-tune a DINOv3~\cite{simeoni2025dinov3} ViT-H/16+ backbone $f_T$ to accept two types of input from each training pair: (i)~the boxed reference $\mathbf{x}_{\text{ref}}^{\text{box}}$, and (ii)~the corresponding detail close-up $\mathbf{x}_{\text{det}}$.
Both views are passed through the same backbone $f_T$ and trained with a shared classification objective over product categories and part types:
\begin{equation}
    \mathcal{L}_{\text{teacher}} = \mathcal{L}_{\text{CE}}(f_{\text{prod}}(f_T(\mathbf{x})), y_{\text{prod}}) + \mathcal{L}_{\text{CE}}(f_{\text{part}}(f_T(\mathbf{x})), y_{\text{part}}), \quad \mathbf{x} \in \{\mathbf{x}_{\text{ref}}^{\text{box}}, \mathbf{x}_{\text{det}}\},
    \label{eq:teacher_loss}
\end{equation}
where $f_{\text{prod}}$ and $f_{\text{part}}$ are simple MLPs classification heads built on the [CLS] token, and $(y_{\text{prod}}, y_{\text{part}})$ is the shared label for each pair.
The crucial design is shared label assignment, assigning the same product$\times$part label to both reference and detail views. It forces the teacher to pull them into the same region of the feature space, completing cross-view semantic alignment entirely within the discriminative model.
After supervised fine-tuning on 80K samples (the total number of reference and detail images), the teacher achieves 84\% joint accuracy, confirming that both views are well-aligned in the shared latent space and ready to serve as distillation targets.

\paragraph{Dual-branch distillation into the generator.}
In the second stage, we distill the aligned teacher representations into both branches of the generator at an intermediate layer $l^*$.
We extract two types of targets from the trained teacher backbone: a semantic [CLS] token $\mathbf{z}_T \in \mathbb{R}^{d_t}$, and patch tokens $\mathbf{Z}_T \in \mathbb{R}^{N_p \times d_t}$ capturing fine-grained spatial structure, where $d_t$ is the output hidden dimension of the teacher.
Let $\mathbf{h}^{(l)}_{\text{text}} \in \mathbb{R}^{L \times D}$ and $\mathbf{h}^{(l)}_{\text{img}} \in \mathbb{R}^{S \times D}$ denote the text-branch and image-branch hidden states at block $l$, respectively.

For the image branch, we provide structural guidance by encoding the ground-truth detail $\mathbf{x}_{\text{det}}$ through the teacher to obtain patch tokens $\mathbf{Z}^{\text{gt}}_T \in \mathbb{R}^{S \times d_t}$.
A lightweight projector $\phi_{\text{img}}: \mathbb{R}^D \to \mathbb{R}^{d_t}$ maps the image-branch hidden states to the teacher space, and the alignment loss maximizes patch-wise cosine similarity following REPA~\cite{yu2025repa}:
\begin{equation}
    \mathcal{L}_{\text{img}} = -\frac{1}{S}\sum_{i=1}^{S} \frac{\phi_{\text{img}}(\mathbf{H}_{\text{img},i}^{(l^*)}) \cdot \mathbf{Z}^{\text{gt}}_{T,i}}{\|\phi_{\text{img}}(\mathbf{H}_{\text{img},i}^{(l^*)})\| \cdot \|\mathbf{Z}^{\text{gt}}_{T,i}\|}.
    \label{eq:img_loss}
\end{equation}
where $S$ is the number of image patch tokens and $\mathbf{H}_{\text{img},i}^{(l^*)}$ denotes the hidden state of the $i$-th patch at layer $l^*$.
For the text branch, we provide semantic guidance by aligning it to the teacher's encoding of the boxed reference $\mathbf{z}^{\text{ref}}_T$.
We mean-pool the text-branch hidden states $\mathbf{H}_{\text{text}}^{(l^*)}$ over non-padding tokens to obtain $\bar{\mathbf{h}}_{\text{text}}$, and minimize the cosine distance through projector $\phi_{\text{text}}: \mathbb{R}^D \to \mathbb{R}^{d_t}$:
\begin{equation}
    \mathcal{L}_{\text{text}} = 1 - \frac{\phi_{\text{text}}(\bar{\mathbf{h}}_{\text{text}}) \cdot \mathbf{z}^{\text{ref}}_T}{\|\phi_{\text{text}}(\bar{\mathbf{h}}_{\text{text}})\| \cdot \|\mathbf{z}^{\text{ref}}_T\|}.
    \label{eq:text_loss}
\end{equation}
The full training loss combines the flow-matching objective with both alignment losses:
\begin{equation}
    \mathcal{L}_{\text{total}} = \mathcal{L}_{\text{flow}} + \lambda_{\text{img}} \cdot \mathcal{L}_{\text{img}} + \lambda_{\text{text}} \cdot \mathcal{L}_{\text{text}},
    \label{eq:total_loss}
\end{equation}
where $\lambda_{\text{img}}$ and $\lambda_{\text{text}}$ are scalar coefficients that balance the distillation strength relative to the primary flow-matching objective.
Because $\mathbf{z}^{\text{ref}}_T$ and $\mathbf{Z}^{\text{gt}}_T$ reside in the same teacher feature space by construction, simultaneously distilling them into the two generative branches causes the branches to converge toward compatible representations, ultimately producing detail images with stronger identity consistency and more precise structural fidelity.

\subsection{Consistency Reward Model and RL Fine-Tuning}
\label{sec:reward}

While CFAD substantially improves cross-branch alignment during supervised training, the generated details may still exhibit subtle deficiencies that pixel-level reconstruction losses cannot capture, such as color drift, implausible fabric textures, or identity deviation from the source garment.
General-purpose preference models~\cite{xu2024imagereward, wu2023hpsv2, ma2025hpsv3} optimize global aesthetics, and editing-specific rewards~\cite{ye2025imgedit, wu2025editreward, luo2025editscore} target instruction-following on natural images; neither directly addresses the consistency requirements unique to fashion detail generation.
We therefore introduce a task-specific \emph{consistency reward model} $R_\psi$ and use it to fine-tune the generator via reinforcement learning.

\paragraph{Multi-dimensional reward formulation.}
Given a boxed reference image and a candidate detail image $(\mathbf{x}_{\text{ref}}^{\text{box}},\, \mathbf{x}_{\text{det}})$, the reward model produces three complementary scores:
The three dimensions are: \textbf{aesthetic plausibility} ($r_{\text{as}}$), which measures visual naturalness and artifact-free quality; \textbf{identity consistency} ($r_{\text{id}}$), which measures whether material, color, pattern, and structure are preserved from the source garment; and \textbf{target fidelity} ($r_{\text{tf}}$), which measures whether the generated close-up correctly corresponds to the indicated local region.
The composite reward is a weighted sum $r = \alpha_{\text{as}}\, r_{\text{as}} + \alpha_{\text{id}}\, r_{\text{id}} + \alpha_{\text{tf}}\, r_{\text{tf}}$.
We fine-tune a vision-language model on a curated dataset of about 60K real reference--detail image pairs with varying quality levels, each annotated with integer scores along the three dimensions. The details of data collection and annotation are provided in Appendix~\ref{app:reward_dataset}.

\paragraph{RL fine-tuning.}
We adopt Negative-aware FineTuning~\cite{diffusionnft2025}, a forward-process RL framework that formulates reward optimization as a single-step policy gradient problem.
Unlike multi-step denoising RL (e.g., DDPO~\cite{black2024ddpo}), NFT operates on a small number of timesteps sampled from the forward process, enabling efficient online optimization.
For each training sample, we generate a group of $K$ candidates, score each with $R_\psi$, and apply group-relative advantage normalization~\cite{shao2024deepseekgrpo} before the policy gradient update.
A small clipping range ensures stable updates on the large-scale base model without catastrophic forgetting of the CFAD-learned alignment.

\section{Experiments}
\label{sec:experiments}

\subsection{FDBench: Task and Data}

\paragraph{Dataset Construction.}
We construct FDBench from a large-scale fashion e-commerce catalog, where each sample is a pair $(\mathbf{x}_{\text{ref}}^{\text{box}},\, \mathbf{x}_{\text{det}})$ consisting of a boxed reference garment image and a corresponding detail close-up. The benchmark spans menswear, womenswear, and childrenswear with a total of 41 product categories. The final benchmark contains $\sim$40K training samples and a held-out 800-sample test set. To keep the main paper concise, we defer full data curation details, including annotation, collection channels, and filtering, to Appendix~\ref{app:fdbench_processing}.



\subsection{Experimental Setup}

\textbf{Evaluation Metrics.}
We evaluate generated detail images along three complementary axes.
SSIM and LPIPS~\cite{zhang2018lpips} measure pixel-level and perceptual similarity between the generated detail and the ground-truth detail when paired ground truth exists.
FID (Fr\'{e}chet Inception Distance) measures distributional similarity between generated and real detail images over the test set. We further report CLIP Image Similarity between the generated detail and the ground-truth detail~\cite{radford2021clip} as CIS, and CLIP Image Similarity between the generated detail and the cropped reference region as CCIS to quantify identity preservation.
We additionally report \textbf{EditScore}~\cite{luo2025editscore} with sub-dimensions PF (Prompt Following), CS (Consistency), PQ (Perceptual Quality), and O (Overall, $\sqrt{\min(\text{PF},\text{CS}) \times \text{PQ}}$); and \textbf{ImageEditJudge}~\cite{ye2025imgedit} with sub-dimensions PD (Physical \& Detail Fidelity), PC (Prompt Compliance), VS (Visual Seamlessness), and the average. Note that for all the metrics related to the Prompt in these two metrics, the text used for the prompt is the text after the PE (Prompt Enhancement) process, in order to obtain an accurate score.


\textbf{Implementation Details.}
We use Qwen-Image-Edit-2511~\cite{qwenimage2025} as our base generator, a 20B-parameter MMDiT with hidden dimension $D=3072$. The DINOv3 ViT-H/16+ view-bridging teacher ($d_t=1280$) is fine-tuned on the 80K FDBench training images with product and part classification heads, achieving 87.2\% part accuracy and 90.6\% product accuracy (84.0\% joint accuracy). For CFAD training, we set the alignment layer $l^*=16$ following REPA~\cite{yu2025repa}, with loss weights $\lambda_{\text{img}}=0.02$ and $\lambda_{\text{text}}=0.05$. The consistency reward model $R_\psi$ is built on Qwen3-VL-8B~\cite{qwen3vl2025}, fine-tuned for 2 epochs on $\sim$85K annotated real reference--detail image pairs with varying quality levels. RL fine-tuning applies DiffusionNFT~\cite{diffusionnft2025} with LoRA (rank$=64$, $\alpha=128$) and composite reward weights $\alpha_{\text{as}}=0.2$, $\alpha_{\text{id}}=0.4$, $\alpha_{\text{tf}}=0.4$.
At inference time, all models generate at $1024\times1024$ resolution with 40 denoising steps; we use classifier-free guidance scale 4.0 for our model and select the recommended guidance scale for each baseline.
Full training hyperparameters are provided in Appendix~\ref{app:impl_details}.

\subsection{Main Results}

We compare \textbf{DetailAnywhere} (Ours) against a comprehensive set of state-of-the-art image editing models, divided into open-source and proprietary categories.
Open-source baselines include Qwen-Image-Edit-2511~\cite{qwenimage2025}, FLUX.2 Dev~\cite{labs2025fluxkontext}, Step1X-Edit~\cite{liu2025step1xedit}, ICEdit~\cite{zhang2025icedit}, FireRed-Image-Edit~\cite{firered2026imageedit}, and JoyAI-Image-Edit~\cite{jd2025joyai}.
Proprietary baselines include GPT-Image-1~\cite{openai2025gptimage}, Nano Banana 2 (Gemini 3.1 Flash Image Preview)~\cite{google2026nanobanana}, and Seedream~5.0 Lite~\cite{bytedance2026seedream}.
For all baselines, the bounding box is rendered on the reference image and a unified instruction: ``Generate a detail view of the boxed region'' is provided as the editing prompt.

Table~\ref{tab:main_results} reports results on the FDBench test set.
DetailAnywhere (CFAD+NFT) achieves the best performance across all five image-quality metrics among open-source methods, substantially outperforming the strongest prompt-enhanced baseline (Qwen-Image-Edit PE) on both pixel-level fidelity (SSIM, FID) and identity preservation (CIS, CCIS).
Notably, baselines without prompt enhancement (Step1X-Edit, ICEdit, FLUX.2 Dev) exhibit extremely low Prompt Following and ImageEditJudge scores despite high Consistency, revealing that they fail to understand the task, and largely copy or minimally modify the reference rather than generating meaningful detail views.
Prompt-enhanced baselines (FireRed PE, Qwen-Image-Edit PE) demonstrate significantly better task comprehension, yet still lag behind DetailAnywhere on identity preservation metrics, confirming that language-level region descriptions alone are insufficient to close the cross-view semantic gap.
Proprietary models benefit from significantly larger model capacity and extensive training corpora, enabling strong task comprehension and high editing-quality scores across the board.
Nevertheless, DetailAnywhere surpasses all three proprietary systems on pixel-level fidelity and identity preservation, and remains competitive on EditScore and ImageEditJudge, validating that our targeted cross-modal alignment strategy can effectively close the gap with large-scale closed-source models.
Comparing the two DetailAnywhere variants, CFAD alone already leads all open-source baselines on ImageEditJudge, and RL fine-tuning with the consistency reward further improves fidelity metrics while maintaining strong editing scores, demonstrating that the dual-branch distillation provides a solid foundation that RL can effectively build upon.

\begin{figure}[t]
  \centering
  \includegraphics[width=\textwidth]{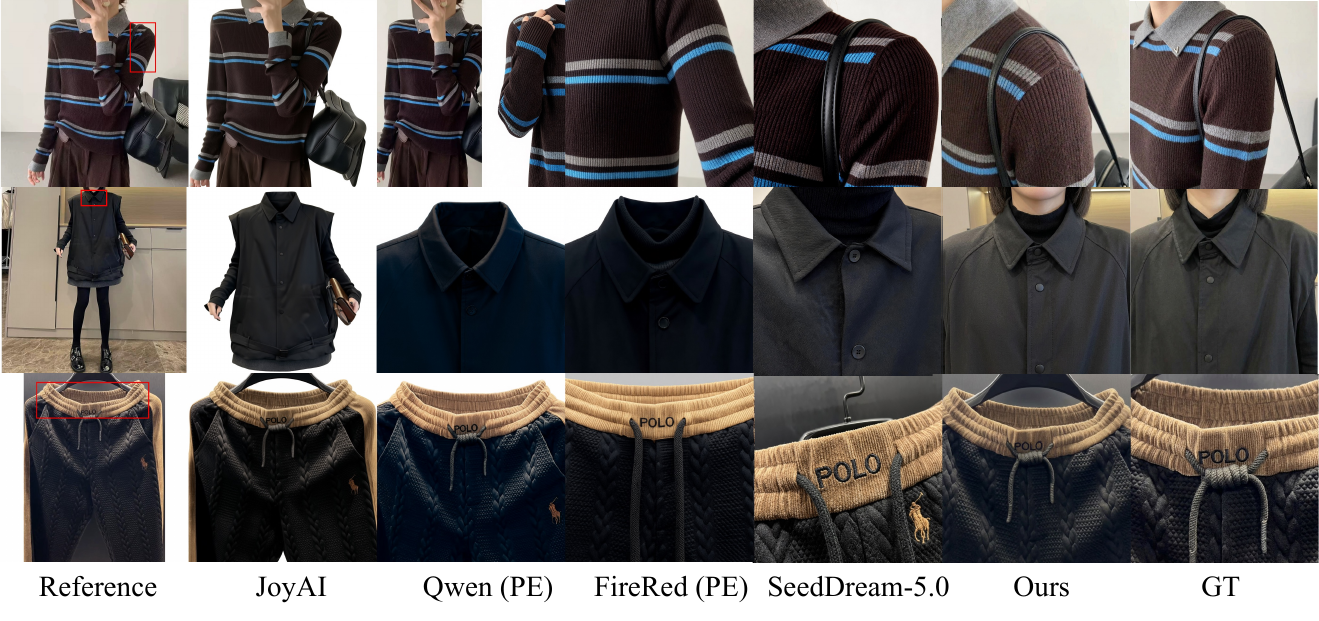}
  \caption{\textbf{Qualitative comparison.} Side-by-side visualization of detail generation results from JoyAI-Image-Edit~\cite{jd2025joyai}, Qwen-Image-Edit (PE)~\cite{qwenimage2025}, FireRed-Image-Edit (PE)~\cite{firered2026imageedit}, Seedream~5.0 Lite~\cite{bytedance2026seedream}, and DetailAnywhere (Ours). Our method produces identity-consistent, photorealistic close-ups with faithful texture and structural preservation.}
  \label{fig:visualization}
\end{figure}

\paragraph{Qualitative analysis.}
Fig.~\ref{fig:visualization} presents visual comparisons on representative FDBench examples.
Our model consistently follows the visual marker and generates the correct garment region, whereas other methods occasionally misidentify the target area or are misled by PE-derived descriptions, as illustrated in the first row.
Beyond localization accuracy, DetailAnywhere also excels in consistency and fidelity: in the second example, only our method faithfully reproduces the smooth button details. In the third example, our model correctly preserves the position of the brand logo, whereas Seedream~5.0 Lite~\cite{bytedance2026seedream} erroneously relocates it to an incorrect position, demonstrating the effectiveness of our consistency reward model in preserving fine-grained spatial details. More visualizations are provided in Appendix~\ref{app:additional_visualizations}.
We further conduct a side-by-side human preference study as a practical indicator for e-commerce deployment; details are provided in Appendix~\ref{app:human_eval}.

\begin{table}[t]
  \caption{\textbf{Main results on FDBench test set.} Comparison of \textbf{DetailAnywhere} against open-source and proprietary baselines. Best in \textbf{bold}, second-best \underline{underlined} (marked within each group). Details of prompt enhancement (PE) are provided in Appendix~\ref{app:prompt_engineering}.}
    \label{tab:main_results}
  \centering
  \resizebox{\textwidth}{!}{%
  \footnotesize
  \begin{tabular}{lcccccccccccccc}
    \toprule
    & \multicolumn{5}{c}{Image Quality} & \multicolumn{4}{c}{EditScore~$\uparrow$} & \multicolumn{4}{c}{ImageEditJudge~$\uparrow$} \\
    \cmidrule(lr){2-6} \cmidrule(lr){7-10} \cmidrule(lr){11-14}
    Method & SSIM~$\uparrow$ & LPIPS~$\downarrow$ & FID~$\downarrow$ & CIS~$\uparrow$ & CCIS~$\uparrow$ & PF & CS & PQ & O & PD & PC & VS & Avg \\
    \midrule
    \multicolumn{14}{l}{\emph{Proprietary}} \\
    GPT-Image-1~\cite{openai2025gptimage}            & 0.274 & 0.650 & 82.6 & 0.797 & 0.726 & 9.16 & 8.82 & \textbf{9.85} & 8.97 & 4.26 & 4.28 & 4.26 & 4.27 \\
    Nano Banana 2~\cite{google2026nanobanana}          & \textbf{0.319} & \textbf{0.619} & \textbf{68.2} & \textbf{0.856} & \underline{0.732} & \textbf{9.78} & \textbf{9.23} & \underline{9.66} & \textbf{9.34} & \underline{4.35} & \underline{4.43} & \underline{4.33} & \underline{4.37} \\
    Seedream 5.0 Lite~\cite{bytedance2026seedream}                & \underline{0.275} & \underline{0.638} & \underline{80.8} & \underline{0.841} & \textbf{0.759} & \underline{9.61} & \underline{9.17} & 9.63 & \underline{9.14} & \textbf{4.46} & \textbf{4.61} & \textbf{4.46} & \textbf{4.51} \\
    \midrule
    \multicolumn{14}{l}{\emph{Open-source}} \\
    FLUX.2 Dev~\cite{labs2025fluxkontext}         & 0.339 & 0.699 & 118.7 & 0.726 & 0.632 & 0.12 & 9.85 & 8.37 & 0.21 & 1.41 & 1.41 & 1.41 & 1.41 \\
    Step1X-Edit~\cite{liu2025step1xedit}             & 0.345 & 0.693 & 120.5 & 0.697 & 0.622 & 1.62 & \textbf{9.55} & 7.77 & 2.32 & 2.97 & 2.97 & 2.96 & 2.97 \\
    ICEdit~\cite{zhang2025icedit}                  & 0.349 & 0.691 & 118.4 & 0.690 & 0.631 & 1.25 & \underline{9.31} & 7.58 & 2.18 & 2.96 & 2.99 & 2.95 & 2.97 \\
    FireRed-Image-Edit~\cite{firered2026imageedit}                & 0.321 & 0.701 & 118.8 & 0.718 & 0.646 & 5.06 & 7.39 & 9.11 & 5.36 & 3.52 & 3.68 & 3.52 & 3.57 \\
    FireRed (PE) & 0.274 & 0.660 & 85.0 & 0.791 & 0.719 & \textbf{9.64} & 8.39 & \underline{9.47} & 8.71 & 3.88 & 3.98 & 3.88 & 3.91 \\
    Qwen-Image-Edit~\cite{qwenimage2025}   & 0.309 & 0.686 & 98.6 & 0.757 & 0.665 & 4.37 & 6.10 & 9.43 & 4.41 & 3.82 & 3.99 & 3.80 & 3.86 \\
    Qwen-Image-Edit (PE) & 0.254 & 0.648 & 82.6 & 0.804 & 0.739 & 9.51 & 8.77 & 9.42 & \underline{8.85} & 4.09 & 4.12 & 4.09 & 4.10 \\
    JoyAI-Image-Edit~\cite{jd2025joyai}       & 0.324 & 0.680 & 93.9 & 0.754 & 0.706 & 7.72 & 8.80 & 8.62 & 7.23 & 3.51 & 3.65 & 3.51 & 3.55 \\
    \midrule
    DetailAnywhere (CFAD)                        & \underline{0.366} & \underline{0.608} & \underline{70.9} & \underline{0.861} & \underline{0.741} & 9.41 & 8.61 & \underline{9.48} & \underline{8.79} & \underline{4.29} & \underline{4.31} & \underline{4.29} & \underline{4.30} \\
    \textbf{DetailAnywhere (CFAD+NFT)} & \textbf{0.394} & \textbf{0.606} & \textbf{66.9} & \textbf{0.863} & \textbf{0.748} & \underline{9.57} & 8.82 & \textbf{9.59} & \textbf{8.95} & \textbf{4.31} & \textbf{4.34} & \textbf{4.30} & \textbf{4.32} \\
    \bottomrule
  \end{tabular}%
  }
\end{table}

\subsection{Ablation Studies}

We conduct two sets of ablations to validate the two key components of DetailAnywhere: the CFAD distillation framework and the consistency reward model.

\paragraph{Ablation on CFAD.}
To isolate the contribution of dual-branch alignment, we compare three variants in Table~\ref{tab:ablation_cfad}: vanilla supervised fine-tuning without any distillation (w/o CFAD), an image-branch-only variant that keeps $\mathcal{L}_{\text{img}}$ but removes text-branch distillation (w/ img-branch), and the full CFAD model with both alignment paths enabled. All variants share the same base model and training schedule without RL fine-tuning.

Full CFAD achieves the best results on the majority of metrics. Adding only image-branch distillation improves pixel-level fidelity (SSIM, FID) and Consistency Score, as the student successfully mimics the teacher's visual encoding of the image~\cite{leng2025repa,yu2025repa}. However, identity preservation metrics fluctuate slightly, because distilling the image branch alone does not help it coordinate with the conditioning branch to understand what to generate. Once the text-branch alignment is introduced, the two branches of the model converge in a shared semantic space, leading to not only the best image-level quality but also a substantial boost in Prompt Following, which reflects that the generator now more correctly interprets the editing instruction. This confirms that image-branch and text-branch distillation are complementary, and both are essential to close the cross-view gap.
A feature distribution analysis in Appendix~\ref{app:feature_distribution} further corroborates this conclusion: CFAD induces clear part-type-aware clustering in the MMDiT's hidden space, providing direct internal evidence for the effectiveness of dual-branch alignment distillation.

\begin{table}[t]
  \caption{\textbf{Ablation on CFAD.} Effect of removing each distillation branch. All variants share the same training schedule and base model, without RL fine-tuning.}
  \label{tab:ablation_cfad}
  \centering
  \footnotesize
  \setlength{\tabcolsep}{4pt}
  \begin{tabular}{lcccccccc}
    \toprule
    & & & & & \multicolumn{4}{c}{EditScore~$\uparrow$} \\
    \cmidrule(lr){6-9}
    Variant & SSIM~$\uparrow$ & FID~$\downarrow$ & CIS~$\uparrow$ & CCIS~$\uparrow$ & PF & CS & PQ & O \\
    \midrule
    w/o CFAD                          & 0.347 & 74.1 & 0.844 & 0.722 & 9.19 & 8.36 & 9.34 & 8.48 \\
    w/ img-branch                   & 0.358 & 71.4 & 0.852 & \textbf{0.743} & 9.21 & 8.56 & 9.42 & 8.56 \\
    \textbf{Full CFAD (Ours)}         & \textbf{0.366} & \textbf{70.9} & \textbf{0.861} & 0.741 & \textbf{9.41} & \textbf{8.61} & \textbf{9.48} & \textbf{8.79} \\
    \bottomrule
  \end{tabular}
\end{table}

\paragraph{Ablation on Reward Model.}
Starting from the same CFAD-trained generator, we compare four RL settings in Table~\ref{tab:ablation_reward}: no RL update, RL with EditReward~\cite{wu2025editreward}, RL with HPSv3~\cite{ma2025hpsv3}, and RL with our task-specific consistency reward from Section~\ref{sec:reward}. All runs use the same DiffusionNFT optimizer, group size, and number of update steps. 
EditReward optimizes general editing quality and improves Perceptual Quality, but substantially degrades pixel-level fidelity and identity preservation, indicating that a generic editing reward misaligns with the consistency requirements of our task.
HPSv3 focuses on global aesthetics and maintains SSIM, yet worsens distributional quality and fails to improve identity metrics.
In contrast, our consistency reward simultaneously improves all fidelity and identity metrics while also boosting EditScore Overall, confirming that a task-specific, multi-dimensional reward grounded in fashion detail consistency is critical for effective RL fine-tuning.

\begin{table}[t]
  \caption{\textbf{Ablation on RL reward.} All variants share the same CFAD-trained initialization and the same DiffusionNFT training schedule; only the reward function differs.}
  \label{tab:ablation_reward}
  \centering
  \footnotesize
  \setlength{\tabcolsep}{4pt}
  \begin{tabular}{lcccccccc}
    \toprule
    & & & & & \multicolumn{4}{c}{EditScore~$\uparrow$} \\
    \cmidrule(lr){6-9}
    Reward & SSIM~$\uparrow$ & FID~$\downarrow$ & CIS~$\uparrow$ & CCIS~$\uparrow$ & PF & CS & PQ & O \\
    \midrule
    No RL                                 & 0.366 & 70.9 & 0.861 & 0.741 & 9.41 & 8.61 & 9.48 & 8.79 \\ 
    EditReward~\cite{wu2025editreward}    & 0.323 & 70.8 & 0.845 & 0.722 & 9.46 & 8.58 & \textbf{9.71} & 8.83 \\
    HPSv3~\cite{ma2025hpsv3}              & 0.365 & 75.1 & 0.851 & 0.736 & 9.17 & 8.65  & 9.66 & 8.62 \\
    \textbf{Consistency Reward (Ours)}    & \textbf{0.394} & \textbf{66.9} & \textbf{0.863} & \textbf{0.748} & \textbf{9.57} & \textbf{8.82} & 9.59 & \textbf{8.95} \\
    \bottomrule
  \end{tabular}
\end{table}


\section{Conclusion}
\label{sec:conclusion}

We formalized fashion detail generation and presented FDBench, the first benchmark comprising 40K+ human-verified reference--detail pairs.
To address the cross-branch semantic gap in MMDiT architectures, we proposed CFAD, which offloads cross-modal alignment to a lightweight view-bridging teacher and distills the aligned representations into both generative branches simultaneously.
A task-specific consistency reward model further improves identity preservation via RL fine-tuning.
Together, these components form \textbf{DetailAnywhere}, which significantly outperforms all open-source baselines and matches state-of-the-art proprietary models across automatic metrics and human evaluations.

\paragraph{Limitations and future work.}
Our evaluation is limited to the fashion domain; generalization to other product categories (e.g., electronics, furniture) remains to be validated.
The view-bridging teacher and the reward model both require paired, annotated data that may not be readily available in all domains.
We plan to extend FDBench to additional domains, explore self-supervised alternatives to the view-bridging teacher, and investigate multi-scale alignment strategies that provide layer-adaptive supervision across the MMDiT depth.

\newpage
\bibliographystyle{unsrtnat}
\bibliography{references}

\newpage
\appendix
\section{Supplementary Materials}

\subsection{Additional Implementation Details}
\label{app:impl_details}
\paragraph{Teacher training.}
The DINOv3 teacher uses a ViT-H/16+ backbone with output dimension $d_t=1280$. We fully fine-tune the backbone with AdamW using a differential learning rate schedule: $2\times10^{-5}$ for the backbone and $1\times10^{-4}$ for the classification heads.
Training uses bf16 mixed precision, batch size 32 per GPU on 4 GPUs, weight decay 0.01, warmup ratio 0.05, label smoothing 0.05, and a maximum of 200 epochs with early stopping on joint accuracy.
The red bounding box rendered on reference images uses a relative line width of 0.06 and an opacity of 0.5.
The two classification losses in Eq.~\eqref{eq:teacher_loss} are weighted as $\omega_{\text{prod}}=0.5$ and $\omega_{\text{part}}=1.0$, reflecting that part-level discrimination is the harder subtask.

\paragraph{CFAD distillation.}
For CFAD training of the base generator, we use DeepSpeed ZeRO-3 with bf16 mixed precision, learning rate $1\times10^{-5}$ with linear warmup over 500 steps, and train for 15 epochs on FDBench ($\sim$40K samples) using 2 GPU nodes with batch size 1 per GPU and gradient checkpointing.
Each distillation projector ($\phi_{\text{img}}$ and $\phi_{\text{text}}$) is a two-layer MLP: LayerNorm$(D)\to$Linear$(D,2048)\to$GELU$\to$Linear$(2048,d_t)$, where $D{=}3072$ and $d_t{=}1280$.

\paragraph{Reward model.}
The consistency reward model $R_\psi$ training uses the same Qwen3-VL-8B~\cite{qwen3vl2025} backbone with learning rate $2\times10^{-5}$ and batch size 64. The annotation dataset comprises $\sim$85K reference-detail pairs with integer scores (1--4) along three dimensions.

\paragraph{RL fine-tuning.}
RL fine-tuning uses DiffusionNFT with LoRA (rank$=64$, $\alpha=128$) on 2 GPU nodes.
We set NFT $\beta=0.1$, PPO clipping range $2\times10^{-4}$, advantage clipping range 5.0, learning rate $8\times10^{-5}$ with AdamW (weight decay $10^{-4}$), and EMA decay 0.9.
Each epoch samples $K{=}16$ candidates per case with per-device batch size 1, 48 unique cases, and 2 gradient steps per epoch.
We sample 2 training timesteps per iteration with time shift 3.0 and generate with 15 inference steps at guidance scale 4.

\subsection{Prompt Enhancement Details}
\label{app:prompt_engineering}
Prompt enhancement (PE) is designed to compensate for the relatively weak region-level understanding of the condition branch in off-the-shelf editing models.
For each boxed reference image, we first use Qwen3-VL-30B-A3B as an external image understanding model to localize the marked region and summarize its garment semantics.
The VLM is prompted to return a strict JSON object with four fields:

\begin{tcolorbox}[colback=gray!5, colframe=gray!60, title={\small\textbf{VLM Understanding Prompt}}, fontupper=\small\ttfamily, breakable]
The image shows a full-view garment, with a red bounding box marking a region.

\medskip
Please carefully inspect the image and answer the following four questions:

1. position: the location of the red box in the image. Choose the most suitable option from: upper-left, top, upper-right, left, center, right, lower-left, bottom, lower-right.

2. clothing\_type: the type of garment, described concisely in no more than 10 words.

3. detail\_part: the garment part corresponding to the red-box region, described concisely in no more than 20 words.

4. textural\_features: distinctive texture, accessory, decoration, or style information of the garment, described concisely in no more than 20 words.

\medskip
Return strict JSON only, without any additional text:

\{"position": "xxx", "clothing\_type": "xxx", "detail\_part": "xxx", "textural\_features": "xxx"\}
\end{tcolorbox}

We then parse the JSON response and assemble the final editing prompt as follows:

\begin{tcolorbox}[colback=blue!3, colframe=blue!40, title={\small\textbf{Assembled Editing Prompt}}, fontupper=\small\ttfamily, breakable]
Generate a high-fidelity close-up detail image of the target garment region. The target region is located at \{position\} in the reference image. The garment is \{clothing\_type\}. The region corresponds to \{detail\_part\}. Preserve the garment identity, including color, material, structure, and the following visual characteristics: \{textural\_features\}. Produce a detail view of this region, with clear texture and faithful local structure.
\end{tcolorbox}

For PE baselines, we remove the red bounding box from the input image and provide the unboxed reference image together with the assembled prompt to the corresponding editing model.
All results marked with ``(PE)'' in Table~\ref{tab:main_results} are obtained using this procedure.

\subsection{Human Preference Evaluation}
\label{app:human_eval}

To complement automatic metrics with practical quality assessment, we conduct a side-by-side (SBS) human preference study on a 800-sample subset of the FDBench test set.
For each test sample, three expert annotators are independently shown the reference image with bounding box together with two generated detail images (Ours vs.\ a baseline) in randomized order, and each annotator selects one of four options: A is better, B is better, both are equally good, or both are equally bad. We will shuffle the order of the images so that the experts cannot tell which one is our result, in order to ensure fairness.
The final label for each sample is determined by majority vote: if at least two of the three annotators agree, the majority option is taken; if all three annotators choose different options, the sample is discarded as inconclusive.
We compare DetailAnywhere against two representative open-source baselines (Qwen-Image-Edit PE and FireRed PE) and the strongest proprietary baseline (Nano Banana 2).
Results are summarized in Table~\ref{tab:human_eval}.

\begin{table}[h]
  \caption{Human preference evaluation (side-by-side). For each pair, annotators label one of four outcomes: Ours wins, Both good, Baseline wins, or Both bad. Percentages are computed over valid samples.}
  \label{tab:human_eval}
  \centering
  \footnotesize
  \begin{tabular}{lcccc}
    \toprule
    Baseline & Ours Wins & Both Good & Baseline Wins & Both Bad \\
    \midrule
    Qwen-Image-Edit (PE) & \textbf{66.1\%} & 2.4\% & 14.9\% & 16.6\% \\
    FireRed (PE)         & \textbf{74.9\%} & 2.0\% & 4.7\%  & 18.3\% \\
    Nano Banana 2        & 40.7\% & 14.5\% & \textbf{41.1\%} & 3.7\%  \\
    \bottomrule
  \end{tabular}
\end{table}

DetailAnywhere is strongly preferred over both open-source PE baselines, winning 66.1\% and 74.9\% of comparisons against Qwen-Image-Edit PE and FireRed PE, respectively, while the baselines are preferred in only 14.9\% and 4.7\% of cases.
Against Nano Banana 2, the win rates are nearly tied (40.7\% vs.\ 41.1\%). This confirms that DetailAnywhere matches the perceived quality of a state-of-the-art proprietary model while decisively outperforming open-source alternatives in human judgment, further validating its readiness for practical e-commerce deployment.

\subsection{FDBench Construction Details}
\label{app:fdbench_processing}

FDBench is constructed through a six-stage pipeline applied to real e-commerce photography, followed by multi-round quality filtering.
The overall flow is summarized below, with all VLM prompts reproduced verbatim.

\paragraph{Stage 1: Reference Image Curation.}
We collect $\sim$85K product-reference image pairs from a large-scale fashion e-commerce platform, selecting only items whose product pages contain at least one dedicated detail photograph.
We first discard images with resolution below 720p, then score the remaining images for aesthetic quality using Doubao-Seed-2.0-pro~\cite{bytedance2026seed}.
Following the evaluation dimensions of HPSv3~\cite{ma2025hpsv3}, we prompt the VLM to assess each image along five axes: composition, color harmony, lighting, texture clarity, and structural integrity.
The scoring prompt is shown below.

\begin{tcolorbox}[colback=gray!5, colframe=gray!60, title={\small\textbf{Aesthetic Scoring Prompt}}, fontupper=\small\ttfamily, breakable]
You are an expert image quality assessor. Evaluate the given product photograph along the following five dimensions, each scored from 1 (worst) to 5 (best):

\medskip
1. Composition: Is the garment well-framed, with a clear subject and balanced layout?

2. Color Harmony: Are the colors natural, well-balanced, and free of color cast?

3. Lighting: Is the lighting even and professional, without harsh shadows or overexposure?

4. Texture Clarity: Are fabric textures, stitching, and material details clearly visible?

5. Structural Integrity: Is the garment properly displayed without wrinkles, distortion, or occlusion?

\medskip
Return strict JSON only:

\{"composition": N, "color\_harmony": N, "lighting": N, "texture\_clarity": N, "structural\_integrity": N, "overall": N\}
\end{tcolorbox}

We retain the top 75\% of images by overall aesthetic score, yielding 60K reference images, each with at least one corresponding real detail photograph.

\paragraph{Stage 2: Detail--Reference Grounding.}
For each existing real detail image, we use Gemini~3.1~Pro~\cite{google2025gemini} to localize the corresponding region on the reference image via bounding-box grounding, producing a spatial annotation for the reference--detail pair.

\begin{tcolorbox}[colback=gray!5, colframe=gray!60, title={\small\textbf{Detail Grounding Prompt}}, fontupper=\small\ttfamily, breakable]
Image 1 is the full product image; Image 2 is a detail reference image. Locate the region in Image 1 that best matches Image 2.

\medskip
Return exactly one bounding box covering only the most matching local region---do not expand to the entire garment.

\medskip
Output a JSON array: [\{"box\_2d": [ymin, xmin, ymax, xmax]\}].

box\_2d uses Image 1 coordinates, range 0--1000.

Do not return markdown, explanations, or any other text.
\end{tcolorbox}

\paragraph{Stage 3: Part-Type Clustering.}
We deploy a locally hosted Qwen3-VL-30B-A3B~\cite{qwen3vl2025} to classify each reference--detail pair into one of 20 predefined garment part types.
The 20 part labels are selected from the top-30 business-relevant detail categories by manual curation, chosen to cover the vast majority of garment structures:
\emph{Collar, Cuff, Fabric/Material, Coat Hem, Pocket, Trouser Cuff, Placket, Bottom Hem, Shoulder Seam, Button, Waistband, Stitching, Skirt Hem, Embroidery, Trouser Waist, Zipper, Quilting, Waistline, Cap Brim, Fur/Wool Texture}.
Note that this taxonomy is intentionally soft: a detail image only needs to contain the relevant part rather than correspond to it exclusively. The part-type label serves solely for dataset balancing and analysis, as it is not used as a generation condition during training or inference.

\begin{tcolorbox}[colback=gray!5, colframe=gray!60, title={\small\textbf{Part-Type Clustering Prompt}}, fontupper=\small\ttfamily, breakable]
You are given two images of a garment. Image 1 is the full product image with a red bounding box marking a region; Image 2 is a close-up detail of that region.

\medskip
Classify the detail region into exactly one of the following 20 part types:

[Collar, Cuff, Fabric/Material, Coat Hem, Pocket, Trouser Cuff, Placket, Bottom Hem, Shoulder Seam, Button, Waistband, Stitching, Skirt Hem, Embroidery, Trouser Waist, Zipper, Quilting, Waistline, Cap Brim, Fur/Wool Texture]

\medskip
If the detail does not include any of the above categories, return ``None''.

\medskip
Return strict JSON only: \{"part\_type": "xxx"\}
\end{tcolorbox}

Pairs classified as \texttt{None} (i.e., unclassifiable) are discarded, leaving 52K grounded reference--detail pairs covering approximately 22K unique reference images.

\paragraph{Stage 4: Automated Quality Filtering.}
We apply a final round of VLM-based quality filtering using Gemini~3.1~Pro, evaluating each pair along three binary dimensions:

\begin{tcolorbox}[colback=gray!5, colframe=gray!60, title={\small\textbf{Quality Filtering Prompt}}, fontupper=\small\ttfamily, breakable]
You will see two images.

Image 1: the original product image (full garment).

Image 2: the candidate detail image.

\medskip
Product category: \{category\}--\{sub\_category\}

Detail part type: \{detail\_part\}

\medskip
Evaluate strictly along three dimensions:

1. part\_match: Does Image 2 actually show a close-up of the specified detail part?

~~- correct: Image 2 clearly shows the specified part in detail.

~~- wrong: Image 2 does not show the specified part, or the part is ambiguous.

2. product\_consistency: Are Image 1 and Image 2 the same product, with consistent details (color, fabric, texture, style)?

~~- correct: Same product, details are consistent.

~~- wrong: Clearly not the same product, or details do not match.

3. structure\_quality: Is the structure of Image 2 reasonable? E.g., garment is complete, no distortion, natural appearance.

~~- correct: Structure is reasonable and natural.

~~- wrong: Obvious structural problems.

\medskip
Return strict JSON only:

\{"part\_match": "correct/wrong", "product\_consistency": "correct/wrong", "structure\_quality": "correct/wrong", "reason": "brief explanation"\}
\end{tcolorbox}

Pairs that fail any of the three checks are removed, reducing the dataset to approximately 46K pairs.

\paragraph{Stage 5: Manual Screening.}
A final round of comprehensive human review is conducted following the same three-dimensional criteria as Stage~4, with additional attention to fine-grained identity preservation: material fidelity, accessory retention (e.g., buttons, zippers, embroidery), and color consistency between the reference and detail images.
After manual screening, the dataset contains \textbf{40893} verified reference--detail pairs.

\paragraph{Stage 6: Train/Test Split.}
We perform stratified random sampling by category proportion to construct an \textbf{800-pair test set} and a \textbf{40093-pair training set}.
The split is conducted at the product-SKU level to prevent data leakage between training and evaluation.

\subsection{FDBench Composition and Split Details}
\label{app:fdbench_composition}

The final FDBench dataset contains \textbf{40{,}893} reference--detail pairs spanning three garment categories and 41 product types.
Table~\ref{tab:category_dist} summarizes the category-level distribution.

\begin{table}[h]
  \caption{FDBench category distribution.}
  \label{tab:category_dist}
  \centering
  \small
  \begin{tabular}{lrr}
    \toprule
    Category & Count & Percentage \\
    \midrule
    Womenswear  & 23{,}389 & 57.2\% \\
    Menswear    &  11{,}577 & 28.3\% \\
    Childrenswear &  5{,}927 &  14.5\% \\
    \midrule
    \textbf{Total} & \textbf{40{,}893} & \textbf{100\%} \\
    \bottomrule
  \end{tabular}
\end{table}

\begin{table}[h]
  \caption{Test set product-type distribution (800 pairs, 32 sub-category types present; the parent category Womenswear/Menswear/Childrenswear is collapsed). The distribution is smoothed relative to the training set to ensure adequate coverage of rare types.}
  \label{tab:test_product_dist}
  \centering
  \small
  \setlength{\tabcolsep}{4pt}
  \begin{tabular}{lr|lr}
    \toprule
    Product Type & Count & Product Type & Count \\
    \midrule
    Short Jacket    & 59 & Skirt           & 24 \\
    Base Layer Top  & 58 & Regular Coat    & 24 \\
    T-shirt         & 53 & Jeans           & 24 \\
    Casual Trousers & 51 & Cardigan        & 21 \\
    Wool Coat       & 40 & Hoodie/Fleece   & 19 \\
    Pullover Knit   & 40 & Sweater         & 19 \\
    Dress           & 37 & Sweatshirt      & 13 \\
    Down Jacket     & 37 & Middle-aged Top & 11 \\
    Jacket          & 32 & Casual Set      & 8 \\
    Shirt           & 29 & Misc.\ Set      & 6 \\
    Padded Jacket   & 29 & Leggings        & 6 \\
    Fur Coat        & 27 & Sportswear Set  & 6 \\
    Casual Pants    & 27 & Padded Coat     & 6 \\
    Denim Trousers  & 27 & Tank Top        & 6 \\
    Vest            & 27 & Leather Jacket  & 5 \\
    Knitwear        & 24 & Leisure Set     & 5 \\
    \midrule
    \multicolumn{2}{c|}{\textbf{Total}} & \multicolumn{2}{c}{\textbf{800}} \\
    \bottomrule
  \end{tabular}
\end{table}

Across product categories, the dataset covers 41 distinct product types, defined as category$\times$sub-category combinations (e.g., Womenswear$\times$T-shirt and Menswear$\times$T-shirt are counted as separate types).
The top-5 product types are: Female Short Jacket (3{,}317), Female Base Layer Top (3{,}040), Female T-shirt (2{,}420), Female Down Jacket (2{,}309), and Female Casual Trousers (2{,}135), while the smallest categories include Female Sweatpants (203) and Male Sportswear Sets (217).

To construct the test set, we apply \emph{smoothed} stratified sampling rather than strict proportional sampling: while the overall category ratio (Womenswear/Menswear/Childrenswear) is approximately preserved, the per-product-type distribution is deliberately flattened so that product types with low frequency in the training set still receive a sufficient number of test samples for reliable evaluation.
This design ensures that the model's generalization to rare product types can be properly assessed.
Table~\ref{tab:test_product_dist} lists the product-type composition of the 800-pair test set.

The split is performed at the product-SKU level to ensure no reference image appears in both sets, preventing data leakage during evaluation.

\subsection{Consistency Reward Dataset and Model Validation}
\label{app:reward_dataset}

\paragraph{Dataset construction.}
We construct the consistency reward dataset from the $\sim$60K reference--detail pairs produced at Stage~2 of the FDBench pipeline (Section~\ref{app:fdbench_processing}), \emph{before} manual screening.
Each pair is re-annotated by Gemini~3.1~Pro using a detailed scoring prompt that evaluates three independent dimensions on a 1--4 integer scale:
\textbf{aesthetic plausibility} (fabric authenticity, craft rationality, visual integrity, commercial usability),
\textbf{product identity consistency} (fabric, craft, trim, and identity-exclusive feature matching against the reference), and
\textbf{target part fidelity} (part accuracy, structure fidelity, detail completeness, focus precision).
Each dimension has strict rubrics with explicit positive/negative indicators; the full scoring prompt is reproduced below.

\begin{tcolorbox}[colback=gray!5, colframe=gray!60, title={\small\textbf{Consistency Reward Scoring Prompt}}, fontupper=\small\ttfamily, breakable]
You are a senior professional judge specializing in fashion product visual quality inspection and cross-image detail consistency verification, with professional knowledge of clothing technology, fabric texture, hardware accessories, and garment structure.

\medskip
\#\#\# Input Information

You will receive 2 fixed inputs, and your evaluation must be strictly based on the following inputs:

1.~~Image A (Reference Image): Full-view product image of the target clothing, with a red bounding box marking the target detail region.

2.~~Image B (Candidate Detail Image): The close-up image to be evaluated, which is supposed to show a detail view of the boxed region in Image A.

\medskip
\#\#\# Evaluation Rules

Evaluate Image B against Image A from the following 3 independent dimensions. For each dimension, you must strictly use only the integer scores from 1 to 4 according to the exclusive scoring criteria for clothing products.

\medskip
---

\#\#\#\# Dimension 1: aesthetic\_plausibility

This dimension focuses on the visual authenticity, physical rationality, and commercial usability of the candidate image as a real clothing product detail shot.

\medskip
Key Criteria:

\textbullet{} Fabric Authenticity: Matches the physical texture and properties of the specified fabric.

\textbullet{} Craft Rationality: Garment structure, stitching and hardware comply with standard clothing production rules.

\textbullet{} Visual Integrity: No artifacts, blurring or distortion that impairs detail recognition.

\textbullet{} Commercial Usability: Meets professional standards for commercial clothing detail imagery.

\medskip
Negative Indicators:

\textbullet{} Severe fabric/structure distortion that is physically impossible.

\textbullet{} Obvious craft defects (twisted stitching, through-modeled hardware) or major visual artifacts.

\textbullet{} Content is unrecognizable as a clothing detail.

\medskip
Scoring Rubric (1-4 Scale):

\textbullet{} 4 (Very Good): Fully meets professional detail shot standards. Fabric, craft and visuals are completely authentic and flawless.

\textbullet{} 3 (Relatively Good): Overall usable and reasonable, with only imperceptible minor flaws that do not affect core detail recognition.

\textbullet{} 2 (Relatively Poor): Has obvious defects in fabric, craft or visual quality that impair usability and detail recognition.

\textbullet{} 1 (Very Poor): Completely unusable. The image is severely distorted, structurally impossible, corrupted or unrecognizable.

\medskip
\#\#\#\# Dimension 2: Product Identity Consistency

This dimension focuses on whether the candidate image belongs to the exact same clothing product as the reference image, verified by matching unique product-specific features.

\medskip
Key Criteria:

\textbullet{} Fabric Consistency: Exact match of fabric material, texture, color and finish to the reference.

\textbullet{} Craft Consistency: Full alignment of stitching, thread details and seam structure with the reference.

\textbullet{} Trim Consistency: Precise match of hardware, buttons, embroidery and decorative parts.

\textbullet{} Identity Exclusivity: Matches unique, non-generic features that distinguish the garment from similar products.

\medskip
Negative Indicators:

\textbullet{} Core product features (fabric, craft, trims) are completely mismatched with the reference.

\textbullet{} Only generic color/category similarity, with no matching of unique product-specific details.

\textbullet{} The candidate image clearly belongs to a different clothing product.

\medskip
Scoring Rubric (1-4 Scale):

\textbullet{} 4 (Very Good): All unique product features are 100\% consistent with the reference, fully confirming the same garment with no deviations.

\textbullet{} 3 (Relatively Good): All core product features are highly matched, with only minor, imperceptible deviations that do not affect same-product judgment.

\textbullet{} 2 (Relatively Poor): Only generic features overlap; core unique features are weakly matched or altered, cannot confirm the same product.

\textbullet{} 1 (Very Poor): Completely unrelated product. All core features contradict the reference, or content is unrecognizable.

\medskip
\#\#\#\# Dimension 3: target\_part\_fidelity

This dimension focuses on how accurately, completely and clearly the candidate image depicts the specified target clothing detail part, while preserving its correct local structure from the reference.

\medskip
Key Criteria:

\textbullet{} Part Accuracy: Content exactly matches the requested target part, with no incorrect parts included.

\textbullet{} Structure Fidelity: Local structure and spatial relationship of the part fully align with the reference.

\textbullet{} Detail Completeness: All core process and decorative details of the target part are fully presented.

\textbullet{} Focus Precision: The target part is the clear focus of the image, with no incorrect framing.

\medskip
Negative Indicators:

\textbullet{} The target part is missing, unrecognizable, or replaced by a wrong part.

\textbullet{} Local structure of the target part is severely distorted or inconsistent with the reference.

\textbullet{} Core details are omitted, or the focus is completely off the requested part.

\medskip
Scoring Rubric (1-4 Scale):

\textbullet{} 4 (Very Good): Accurately and completely depicts the target part, with structure, details and framing perfectly consistent with the reference.

\textbullet{} 3 (Relatively Good): Clearly presents the target part with all core details matched, with only minor framing/focus deviations that do not affect recognition.

\textbullet{} 2 (Relatively Poor): The target part is present but has obvious deviations in structure, missing details, or incorrect framing.

\textbullet{} 1 (Very Poor): Completely fails to show the target part. The part is wrong, missing, unrecognizable, or the image is corrupted.

\medskip
\#\#\# Mandatory Scoring Constraints (must be strictly followed)

1.~~Independent Judgment: The three dimensions must be scored independently, and the score of one dimension must not interfere with the others.

2.~~High Score Restriction: Generic similarity is not enough for a score of 3 and above. High scores must have clear and verifiable matching evidence of exclusive clothing details.

3.~~Strict Principle for Uncertainty: When there is any uncertainty about feature matching, part accuracy, or rationality, you must choose the lower score gear, and overestimation is strictly prohibited.

4.~~Score Range Limit: Only integer scores from 1 to 4 are allowed.

\medskip
\#\#\# Output Requirements

Return only a single standard JSON object without any additional text, notes, or explanations, strictly following the schema below. The "reason" must be a specific description corresponding to the scoring criteria, pointing out the exact clothing detail features/mismatches/defects, and vague descriptions such as "has defects" or "good consistency" are prohibited.

\medskip
\{"aesthetic\_plausibility": \{"score": 1-4, "reason": "specific reason corresponding to the scoring criteria"\}, "product\_identity\_consistency": \{"score": 1-4, "reason": "specific reason corresponding to the scoring criteria"\}, "target\_part\_fidelity": \{"score": 1-4, "reason": "specific reason corresponding to the scoring criteria"\}\}
\end{tcolorbox}

We retain only the three integer scores from each response, discarding the free-text reasons, to form the final \textbf{$\sim$60K consistency reward dataset}.

\paragraph{Model training.}
We fine-tune Qwen3-VL-8B~\cite{qwen3vl2025} as the reward backbone via supervised fine-tuning for 2 epochs. The model takes a boxed reference image (with the red bounding box rendered) and the candidate detail image as input, without any text description from prompt enhancement.
The input prompt is fixed as:

\begin{tcolorbox}[colback=gray!5, colframe=gray!60, title={\small\textbf{Reward Model Input Prompt}}, fontupper=\small\ttfamily]
<image> is the reference image with a red bounding box marking the target region. <image> is a candidate detail image of the boxed region. Please score the image pair along aesthetic\_plausibility, Product Identity Consistency, and target\_part\_fidelity. For each dimension, you must strictly use only the integer scores from 1 to 4 according to the exclusive scoring criteria for clothing products.
\end{tcolorbox}

\paragraph{Validation.}
We evaluate the trained reward model on a held-out test set of 200 reference--detail pairs (100 positive, 100 negative).
Positive samples are pairs that passed the Stage~5 manual screening in the FDBench pipeline; negative samples are pairs that were rejected during the same screening round.
Table~\ref{tab:reward_validation} reports the mean normalized score (0--1 scale) for positive and negative samples, their separation (gap), effect size (Cohen's $d$), and statistical significance (Welch's $t$-test).

\begin{table}[h]
  \caption{Consistency reward model validation on 200 held-out pairs (100 positive, 100 negative). Scores are normalized to $[0,1]$. $^{***}$: $p < 0.001$.}
  \label{tab:reward_validation}
  \centering
  \small
  \begin{tabular}{lcccc}
    \toprule
    Dimension & Positive $\mu$ & Negative $\mu$ & Gap & Cohen's $d$ \\
    \midrule
    Aesthetic Plausibility        & 0.923 & 0.907 & 0.017 & 0.08 \\
    Identity Consistency$^{***}$  & 0.760 & 0.547 & 0.213 & 0.54 \\
    Target Fidelity$^{***}$       & 0.833 & 0.617 & 0.217 & 0.60 \\
    \bottomrule
  \end{tabular}
\end{table}

The reward model achieves statistically significant separation on the two consistency-critical dimensions: identity consistency ($p=0.0002$, $d=0.54$) and target fidelity ($p<0.0001$, $d=0.60$), confirming its ability to discriminate between high-quality and defective detail generations.
The aesthetic plausibility dimension shows a small, non-significant gap ($d=0.08$), which is expected: rejected pairs in Stage~6 were primarily filtered for identity or fidelity failures rather than aesthetic defects, so both positive and negative samples tend to have comparable visual quality.

\subsection{Feature Distribution Analysis of CFAD}
\label{app:feature_distribution}

To directly verify that CFAD aligns the internal representations of the MMDiT, we extract hidden features from four layers (10, 16, 20, 30) of the image branch and text branch for both the CFAD-trained model and a vanilla SFT baseline (trained with the same schedule but without any distillation loss).
We select 90 test samples (30 per category) from three representative part types---Cuff, Collar, and Pocket---and perform a single forward pass per sample at a fixed intermediate timestep ($t \approx 0.5$) to capture the mean-pooled hidden states.

We measure intra-class cosine similarity (average pairwise similarity among samples of the same part type) and inter-class cosine similarity (average similarity between samples of different part types), and report their difference as the \emph{separation} score.
Higher separation indicates that the features form tighter, more discriminative clusters by part type.
Results are summarized in Table~\ref{tab:feature_cluster}.

\begin{table}[h]
  \caption{\textbf{Feature clustering analysis across MMDiT layers.} Intra-/inter-class cosine similarity and separation score for the image branch and text branch at layers 10, 16, and 30. CFAD substantially improves separation in both branches, with layer~16 (the alignment target $l^*$) showing the largest gain.}
  \label{tab:feature_cluster}
  \centering
  \small
  \begin{tabular}{clcc}
    \toprule
    Layer & Model & Img-Branch Sep. & Text-Branch Sep. \\
    \midrule
    \multirow{2}{*}{10} & w/o CFAD & 0.00030 & 0.00086 \\
                        & w/ CFAD  & 0.00191 & 0.00624 \\
    \midrule
    \multirow{2}{*}{16} & w/o CFAD & 0.00056 & 0.00062 \\
                        & w/ CFAD  & 0.00521 & 0.03638 \\
    \midrule
    \multirow{2}{*}{20} & w/o CFAD & 0.00017 & 0.00039 \\
                        & w/ CFAD  & 0.00143 & 0.01469 \\
    \midrule
    \multirow{2}{*}{30} & w/o CFAD & 0.00006 & 0.00034 \\
                        & w/ CFAD  & 0.00033 & 0.00382 \\
    \bottomrule
  \end{tabular}
\end{table}

\begin{figure}[h]
  \centering
  \includegraphics[width=0.65\linewidth]{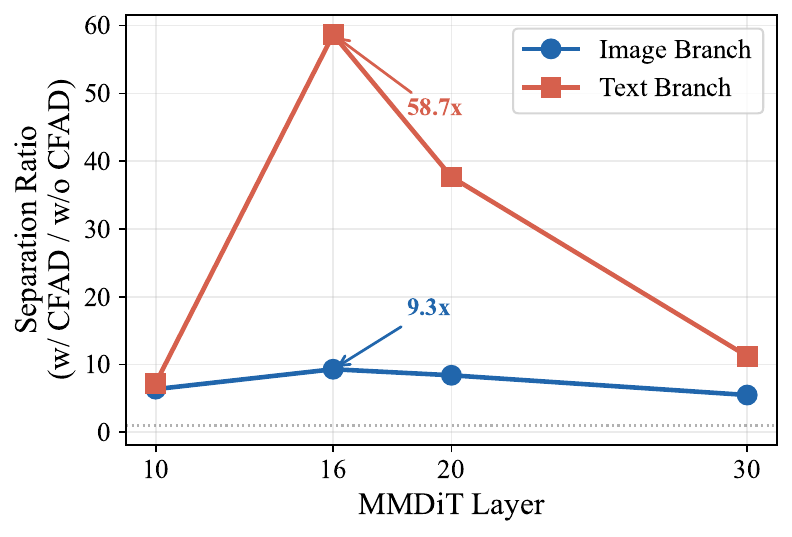}
  \caption{\textbf{Separation ratio across MMDiT layers.} Ratio of the separation score (w/ CFAD / w/o CFAD) for image and text branches. Both branches peak at the alignment target layer~16 ($l^*$), with the text branch reaching $58.7\times$.}
  \label{fig:separation_across_layers}
\end{figure}

Without CFAD, all layers exhibit negligible separation ($<$0.001) in both branches: the features are essentially collapsed into an undifferentiated cluster regardless of which garment region the bounding box indicates.
To better illustrate the layer-wise effect, Fig.~\ref{fig:separation_across_layers} plots the \emph{separation ratio} (w/ CFAD divided by w/o CFAD) across the four layers.
Both branches peak sharply at the alignment target layer~16: the text branch achieves a $58.7\times$ ratio, and the image branch a $9.3\times$ ratio.
Neighboring layers also benefit, with layer~20 still showing $37.7\times$ for text and $8.4\times$ for image, but the gain attenuates rapidly toward earlier and later layers.
This peaked pattern is consistent with the REPA design~\cite{yu2025repa}: distillation is applied at a single intermediate layer, and the alignment signal propagates partially to adjacent layers but is strongest at the injection point.
The disproportionately large text-branch improvement confirms that text-branch distillation is critical for injecting region-level semantic understanding, which is precisely the bottleneck identified in Section~\ref{sec:cfad}.
These results provide direct evidence that CFAD successfully induces part-type-aware clustering in the MMDiT's internal feature space, corroborating the quantitative gains observed in the main ablation study (Table~\ref{tab:ablation_cfad}).

\subsection{Additional Visualizations}
\label{app:additional_visualizations}
\begin{figure}[H]
  \centering
  \includegraphics[width=0.90\linewidth]{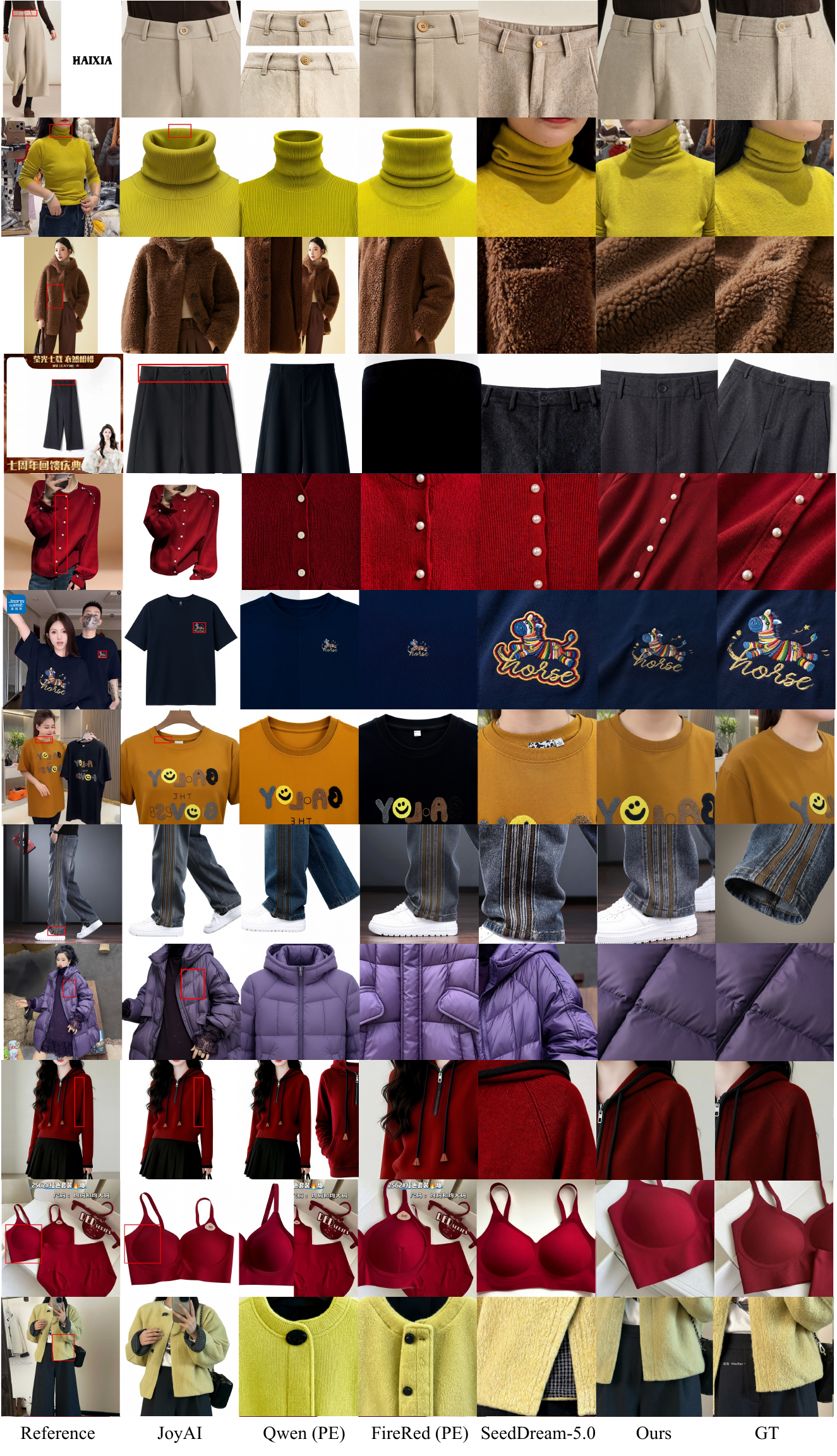}
  \caption{\textbf{Additional Fashion Detail Visualizations.} Additional visualizations.}
  \label{fig:additional_visualizations}
\end{figure}

In addition to the main paper visualization, we provide more side-by-side comparisons of generated detail images from different methods on representative FDBench examples in Fig.~\ref{fig:additional_visualizations}.

\end{document}